\documentclass[runningheads]{llncs}

\usepackage{amsmath, amssymb}
\usepackage{algorithm}
\usepackage{algorithmic}

\usepackage[mobile]{eccv}

\usepackage{eccvabbrv}

\usepackage{graphicx}
\usepackage{booktabs}
\usepackage{multirow}
\usepackage{makecell}
\usepackage{booktabs}
\usepackage{multirow}
\usepackage{threeparttable}
\usepackage{tabularx}      
\usepackage{siunitx}       
\usepackage[table]{xcolor} 
\usepackage{makecell}      
\usepackage{arydshln}
\usepackage[accsupp]{axessibility}  

\usepackage{hyperref}

\usepackage{orcidlink}

\begin{document}

\title{FDIF: Formula-Driven Supervised Learning \\ with Implicit Functions \\for 3D Medical Image Segmentation} 

\titlerunning{FDIF: Formula-Driven Supervised Learning with Implicit Functions}

\author{Yukinori Yamamoto\inst{1, 3}\thanks{\email{yamanokoai@gmail.com}} \and Kazuya Nishimura\inst{2} \and Tsukasa Fukusato\inst{1} \and\\ Hirokazu Nosato\inst{3} \and Tetsuya Ogata\inst{1, 3} \and Hirokatsu Kataoka\inst{3, 4}}

\authorrunning{Y.~Yamamoto et al.}

\institute{Waseda University \and The University of Osaka \and National Institute of Advanced Industrial Science and Technology \and University of Oxford}

\maketitle

\begin{abstract}
Deep learning-based 3D medical image segmentation methods relies on large-scale labeled datasets, yet acquiring such data is difficult due to privacy constraints and the high cost of expert annotation. Formula-Driven Supervised Learning (FDSL) offers an appealing alternative by generating training data and labels directly from mathematical formulas. However, existing voxel-based approaches are limited in geometric expressiveness and cannot synthesize realistic textures.
We introduce Formula-Driven supervised learning with Implicit Functions (FDIF), a framework that enables scalable pre-training without using any real data and medical expert annotations. FDIF introduces an implicit-function representation based on signed distance functions (SDFs), enabling compact modeling of complex geometries while exploiting the surface representation of SDFs to support controllable synthesis of both geometric and intensity textures.
Across three medical image segmentation benchmarks (AMOS, ACDC, and KiTS) and three architectures (SwinUNETR, nnUNet ResEnc-L, and nnUNet Primus-M), FDIF consistently improves over a formula-driven method, and achieves performance comparable to self-supervised approaches pre-trained on large-scale real datasets. We further show that FDIF pre-training also benefits 3D classification tasks, highlighting implicit-function-based formula supervision as a promising paradigm for data-free representation learning.
Code is available at \url{https://github.com/yamanoko/FDIF}.

\end{abstract}

\section{Introduction}
\label{sec:introduction}

The scarcity of labeled data remains a critical bottleneck for deep learning-based 3D medical image analysis. Deep learning models have substantially improved the localization of tumors and organs in CT and MRI, contributing to more accurate diagnosis and treatment planning~\cite{milletari2016v,isensee2021nnu,hatamizadeh2021swin}. However, achieving strong performance typically requires large-scale labeled datasets, which are difficult to obtain in medical imaging due to strict privacy regulations that limit data sharing and the high cost of expert annotation for 3D volumetric data.

To address this challenge, self-supervised learning (SSL) has emerged as a paradigm for learning representations from unlabeled data. For example, Wald~\etal~\cite{wald2025openmind} compiled a dataset of 114K 3D brain MRI volumes from over 800 sources and demonstrated that SSL pre-training improves downstream 3D medical segmentation.
While SSL has proven highly effective and continues to advance the field, it still presents several practical limitations in the context of 3D medical segmentation. First, although SSL removes the need for labeled data, it still requires large collections of unlabeled medical images for pre-training, which are still difficult to acquire and share at scale due to privacy regulations and data governance constraints. Second, in encoder--decoder architectures commonly used for segmentation, SSL typically pre-trains only the encoder, leaving the decoder randomly initialized because pretext tasks do not involve pixel-wise prediction.

Formula-Driven Supervised Learning (FDSL)~\cite{Kataoka2021-ns,kataoka2025pretrainingvisiontransformersformuladriven,9878798} generates synthetic data (i.e., images and labels) from mathematical formulas, enabling task-consistent pre-training without real data and alleviating privacy concerns associated with sensitive datasets such as medical images.
PrimGeoSeg~\cite{tadokoro2024primitive} applied this idea to 3D medical segmentation by constructing synthetic objects from geometric primitives and assigning segmentation labels to each component. Despite using no real data, it achieved performance comparable to SSL-based approaches and synthetic data generation methods that utilize real medical images~\cite{dey2024learninggeneralpurposebiomedicalvolume}.

\begin{figure}[t]
\centering
\includegraphics[width=\textwidth]{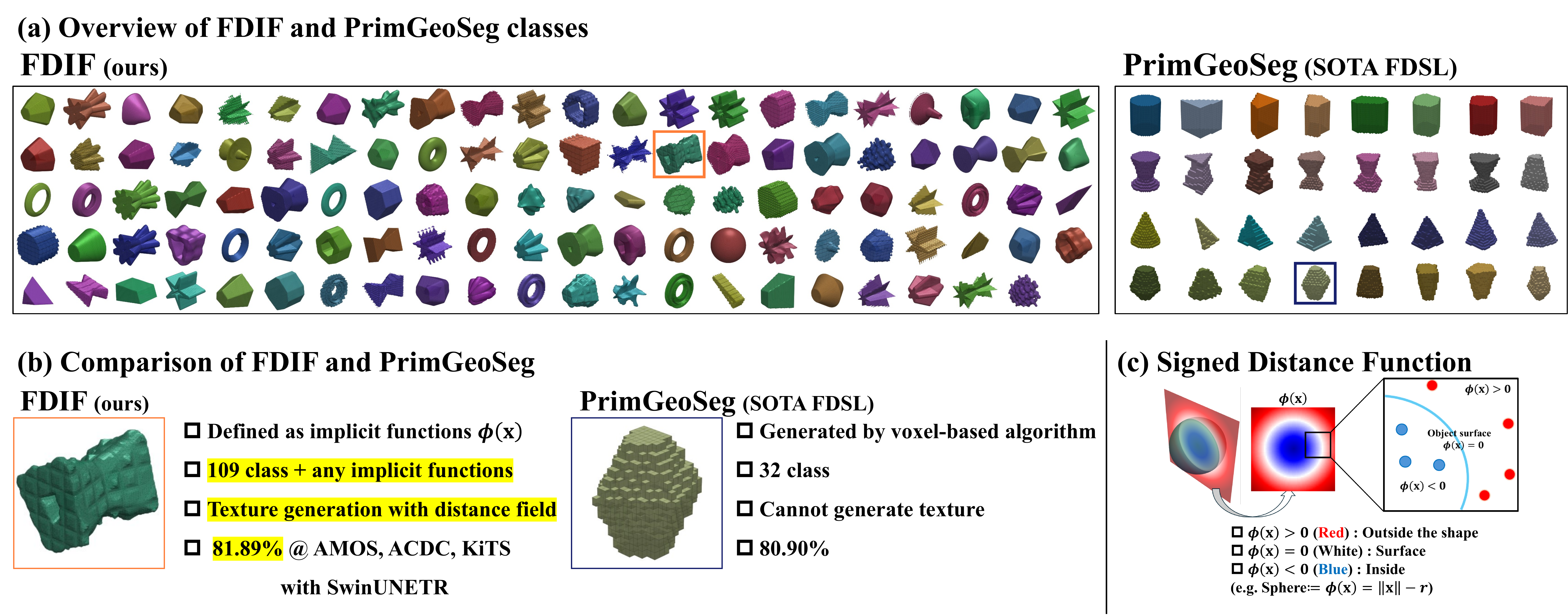}
\caption{(a) Overview of FDIF (109 classes) and PrimGeoSeg (32 classes). (b) Comparison of FDIF and PrimGeoSeg. FDIF uses implicit functions and leverages it for texture generation. (c) A signed distance function (SDF) reflects distance to the  nearest surface: positive outside, negative inside, and zero on the boundary.}
\label{fig:intro}
\end{figure}

However, PrimGeoSeg has two main limitations. First, shapes are generated by extruding 2D cross-sections, which restricts the range of representable geometries and makes it difficult to express complex topologies such as holes or cavities (in Fig. \ref{fig:intro} (a)). Second, voxel grids lack explicit object boundary information, making it difficult to synthesize realistic intensities or textures that depend on surface and depth structure. These limitations stem from the discrete nature of voxel representations, which do not provide a globally consistent description of shape boundaries. This issue is particularly critical in medical imaging, where accurately capturing diverse anatomical shapes, intra-object heterogeneity, and boundary contrast is essential for medical image recognition.

To address these limitations, we propose \textit{Formula-Driven supervised learning with Implicit Functions} (\textbf{FDIF}), a framework that represents synthetic objects using signed distance functions (SDFs). By modeling shapes as continuous 3D functions, SDFs enable flexible generation of diverse geometric variations.
A key advantage of SDFs is that they explicitly encode the distance to the object surface, naturally providing surface information that can be exploited to control appearance (Fig.~\ref{fig:intro} (c)). Building on this property, FDIF introduces two mechanisms: (1) a displacement function that perturbs the distance field to generate geometric surface textures, and (2) a surface-driven intensity mapper that assigns voxel intensities based on the distance to the boundary. This design enables controlled generation of both geometric and appearance variations.
In this work, we use simple procedural functions to build the shape, displacement, and mapper libraries, ensuring geometric diversity and clear inter-class separability. This enables the generation of synthetic volumes with flexible shapes, textured surfaces, and structured intensity patterns, which are key cues for medical image segmentation. The formulation is in principle extensible by introducing new SDFs and scalar functions.

Extensive experiments demonstrate the effectiveness of FDIF. Across three segmentation benchmarks and three architectures, FDIF consistently outperforms both training from scratch and PrimGeoSeg in average Dice score, and achieves performance comparable to SSL methods pre-trained on large-scale real data despite using no real data. Furthermore, extending FDIF to 3D classification tasks shows that implicit-function-based pre-training generalizes beyond segmentation to broader 3D recognition tasks.

Our contributions are summarized as follows:
\begin{itemize}
\item We propose \textbf{Formula-Driven supervised learning with Implicit Functions (FDIF)}, a framework that leverages signed distance functions (SDFs) to generate diverse synthetic labeled volumes for supervised pre-training in 3D medical image segmentation without using real data.

\item We introduce an extensible SDF-based library that enables flexible generation of diverse geometries and appearance variations. The library supports geometric and intensity texture synthesis through displacement and surface-driven intensity mapping based on the signed distance field.

\item Extensive experiments show that FDIF consistently outperforms the SOTA formula-driven method while achieving performance comparable to self-supervised methods trained on real data. FDIF also improves performance on 3D classification tasks.
\end{itemize}

\section{Related Work}
\label{sec:related_work}

\subsection{Pre-training for 3D Medical Image Segmentation}
\label{sec:rw_ssl}

Self-supervised learning (SSL) has become the dominant pre-training paradigm for 3D medical image analysis, with numerous methods proposed including contrastive learning~\cite{zhou2021models,xie2022unimiss,jiang2022self,wu2024voco,wang2023mis,tang2022selfsupervisedpretrainingswintransformers} and masked image modeling~\cite{chen2023masked,zhuang2025advancing}.
Wald~\etal~\cite{Wald2024-ds} revisit masked autoencoders (MAE) specifically for 3D CNNs, showing that a properly optimized MAE with a ResEnc U-Net architecture and 39K brain MRI volumes surpasses prior SSL methods by approximately 3 Dice points.
Building on this finding, Wald~\etal~\cite{wald2025openmind} publish the largest publicly available 3D pre-training dataset (114K brain MRI volumes) and benchmark seven SSL methods across CNN and Transformer architectures, establishing current best practices for 3D SSL pre-training.
Xu~\etal~\cite{xu2025generalizable} further scale SSL by adapting DINO to 3D and pre-training on approximately 100K multi-organ, multi-modality scans, achieving state-of-the-art transfer performance across diverse downstream tasks.

While these advances are impressive, SSL methods rely on large-scale unlabeled datasets, which are difficult to collect in medical imaging due to privacy regulations and specialized acquisition protocols.

\subsection{Formula-Driven Supervised Learning}
\label{sec:rw_fdsl}

Formula-Driven Supervised Learning (FDSL)~\cite{Kataoka2021-ns} offers a fundamentally different approach: both training images and their labels are generated algorithmically from mathematical formulas, eliminating the need for real data entirely. The original FDSL work constructs FractalDB, a database of fractal images with automatically assigned category labels, and shows that CNNs pre-trained on FractalDB can partially match the accuracy of ImageNet-pre-trained models.
Kataoka~\etal~\cite{kataoka2025pretrainingvisiontransformersformuladriven} extend FDSL to Vision Transformers, showing that ExFractalDB-21K achieves 83.8\% top-1 accuracy on ImageNet-1k after fine-tuning, approaching JFT-300M-level performance with $14.2\times$ fewer images and without using real data.
Yamada~\etal~\cite{9878798} extend FDSL to the 3D domain by constructing PC-FractalDB, a point-cloud fractal dataset that leverages natural 3D fractal structures for pre-training 3D object detection models.
PrimGeoSeg~\cite{tadokoro2024primitive} further adapts FDSL to 3D medical image segmentation. It generates synthetic labeled volumes by composing geometric primitives using 8 cross-sectional rules and 4 extrusion rules, producing 32 shape classes for supervised pre-training with the standard segmentation loss. Remarkably, PrimGeoSeg achieves performance comparable to or exceeding SSL methods despite using no real data.

However, voxel-based representations restrict shapes to extruded structures and cannot model boundary-aware textures, which our SDF-based method resolves.

\subsection{Implicit Function Representations}
\label{sec:rw_implicit}

Signed distance functions (SDFs) are a classical implicit representation that assigns to each spatial point its signed distance to the nearest surface boundary~\cite{10.1145/37402.37422}.
In computer vision and graphics, SDFs have been widely used for surface reconstruction, shape modeling, and rendering, thanks to their ability to represent complex topologies compactly and support efficient geometric operations such as Boolean combinations, offsetting, and smooth blending \cite{park2019deepsdf,mildenhall2020nerf,sitzmann2020implicit,NEURIPS2020_55053683}.
Recent advances in deep learning have greatly expanded the role of implicit functions in 3D vision.
Park~\etal~\cite{park2019deepsdf} propose DeepSDF, which learns a continuous SDF conditioned on a latent code, enabling high-quality shape representation, interpolation, and completion for entire object classes from partial or noisy 3D input.
Mildenhall~\etal~\cite{mildenhall2020nerf} introduce Neural Radiance Fields (NeRF), which represents a scene as a continuous volumetric function mapping 5D coordinates (spatial location and viewing direction) to volume density and radiance, achieving photorealistic novel-view synthesis via differentiable volume rendering.
These works demonstrate the power of implicit representations for \emph{reconstructing} or \emph{rendering} 3D scenes from observations.

However, prior work has not explored their role as a mechanism for generating structured supervision signals for representation learning.
We leverage SDF properties to generate diverse labeled 3D training volumes, enabling compact shape construction and natural geometric and intensity textures via simple distance-field transformations.

\section{Method}
\label{sec:method}

We propose Formula-Driven supervised learning with Implicit Functions (FDIF), 
a framework for generating diverse synthetic 3D labeled volumes for supervised pre-training using signed distance functions (SDFs). 
This approach overcomes the limitations of conventional voxel-based data generation \cite{tadokoro2024primitive}, 
which is restricted to extruded shapes and lacks a globally consistent distance field to object boundaries (Fig.~\ref{fig:intro} (b)). 
By contrast, SDFs provide a continuous distance field defined over $\mathbb{R}^3$, 
enabling faithful geometric modeling, surface perturbation, and function-based intensity generation (Fig.~\ref{fig:intro} (c)).

\noindent
{\bf Overview of the Proposed Method.}
An overview of the framework is shown in Algorithm~\ref{alg:data_generation}. 
Our goal is to generate a dataset 
\[
\mathcal{D} = \{(\mathbf{I}_i, \mathbf{Y}_i)\}_{i=1}^{N},
\]
where $\mathbf{I}_i \in \mathbb{R}^{H \times W \times D}$ denotes a synthetic 3D image and 
$\mathbf{Y}_i$ its corresponding segmentation mask. 
The dataset is used to pre-train 3D medical image segmentation networks such as nnU-Net.

\begin{figure}[!t]
\centering
\includegraphics[width=\textwidth]{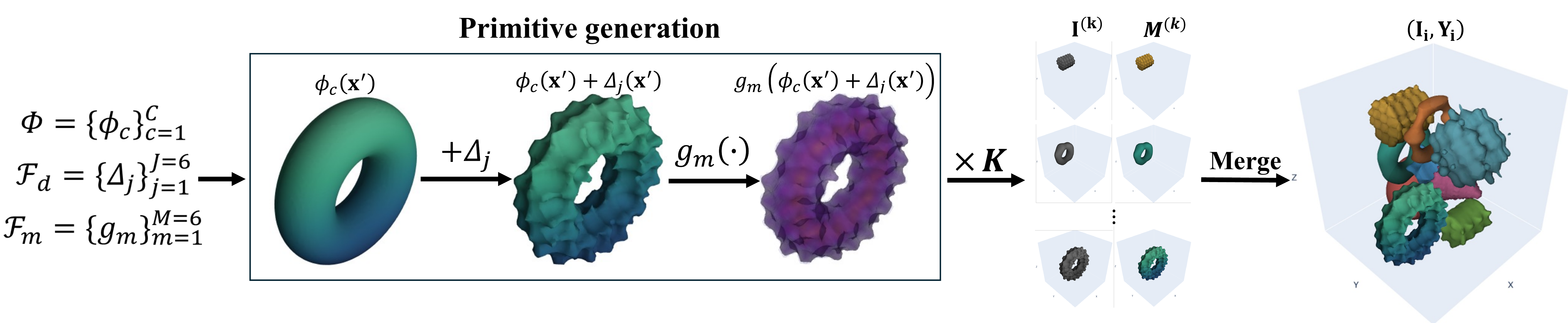}
\caption{Synthetic volume generation via primitive composition. Each primitive is assigned a base SDF from a diverse SDF library $\Phi$, transformed with random spatial parameters, augmented with displacement $\Delta_j$ for geometric texture, and converted to intensity via mapper functions $g_m$. Multiple primitives are merged to form the final labeled volume.}
\label{fig:data_generation_flow}
\end{figure}

The framework consists of two main stages:
1) construction of function libraries (Sec.~\ref{sec:method_library}), 
2) synthetic volume generation via primitive composition (Sec.~\ref{sec:data_gene} and Fig.~\ref{fig:data_generation_flow}).

\subsection{Construction of Function Libraries}
\label{sec:method_library}

This section describes the construction of the function libraries used for synthetic data generation: the Signed Distance Function (SDF) library, the Displacement Function (DF) library, and the Mapper Function (MF) library.

We first construct an SDF library composed of a collection of signed distance functions representing base shapes. Since an SDF encodes the distance from the surface at every spatial location, it provides a natural foundation for structured transformations. Building on this property, we introduce a DF library to generate geometric textures by perturbing distance fields, and an MF library to produce intensity patterns by mapping signed distance values to voxel intensities.

\noindent
{\bf Construction of the SDF library.}
A signed distance function (SDF) is defined as 
$\phi : \mathbb{R}^3 \rightarrow \mathbb{R}$,
which assigns to each point $\mathbf{x}$ its signed Euclidean distance to the closest surface boundary:
\[
\phi(\mathbf{x}) < 0 \; (\text{inside}), \quad
\phi(\mathbf{x}) = 0 \; (\text{boundary}), \quad
\phi(\mathbf{x}) > 0 \; (\text{outside}).
\]

We construct a library of SDFs
\[
\Phi = \{\phi_c\}_{c=1}^{C},
\]
where each $\phi_c$ defines a distinct 3D object via its zero level set 
$\{\mathbf{x} \mid \phi_c(\mathbf{x}) = 0\}$.

The pool contains $C=109$ classes.
Three primitive solids (sphere, octahedron, and cone) are included.
The remaining 106 classes are generated through geometric operations such as extrusion, revolution, and hollowing applied to 2D base shapes.
The library is designed to ensure geometric diversity and inter-class separability.

All SDFs are defined in closed form (see Supplementary Material), enabling exact surface representation and efficient sampling.

\noindent
{\bf DF Library.}
Object texture is important for interpreting medical images because many anatomical structures are defined not only by their overall shape but also by local geometric patterns (e.g., tumor margins, vessel wall layers, trabecular bone). To model these variations during pretraining, a displacement-function library is introduced that perturbs the base signed distance field (SDF), allowing the generated shapes to include realistic geometric textures.

We construct a displacement-function library $\mathcal{F}_d$ consisting of six parametric displacement families:
\begin{equation}
\mathcal{F}_d
=
\{\Delta_1, \Delta_2, \Delta_3, \Delta_4, \Delta_5, \Delta_6\}.
\end{equation}

\begin{figure}[t]
\centering

\begin{minipage}[t]{0.49\linewidth}
\centering
\includegraphics[width=0.95\linewidth]{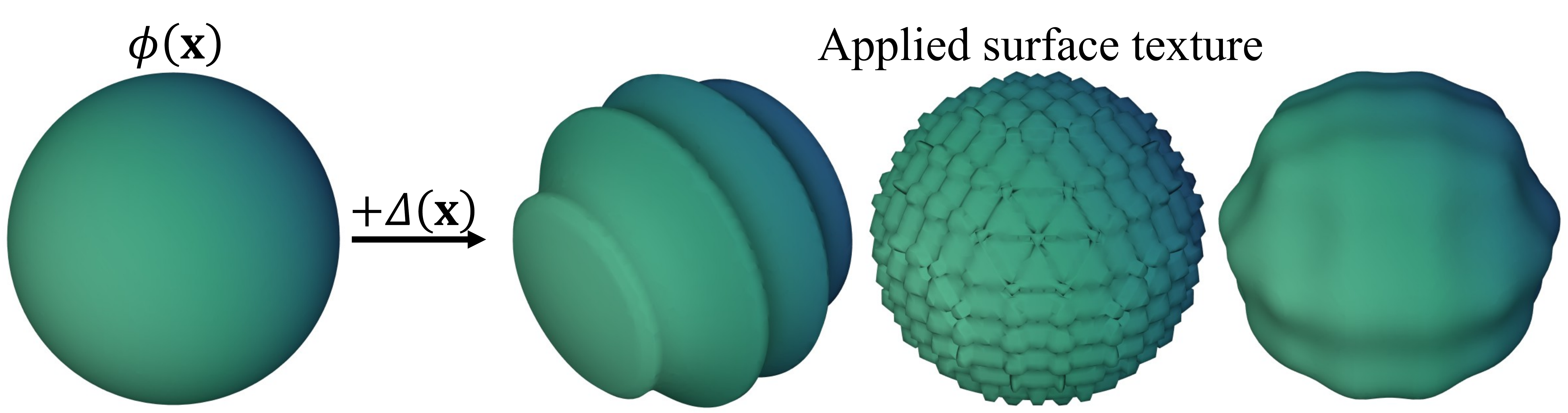}
\caption{Additive displacement on an SDF.
Left: base sphere $\phi(\mathbf{x})$.
Different displacement functions $\Delta(\mathbf{x})$ generate diverse textures while preserving a closed surface.}
\label{fig:displacement_comparison}
\end{minipage}
\hfill
\begin{minipage}[t]{0.49\linewidth}
\centering
\includegraphics[width=\linewidth]{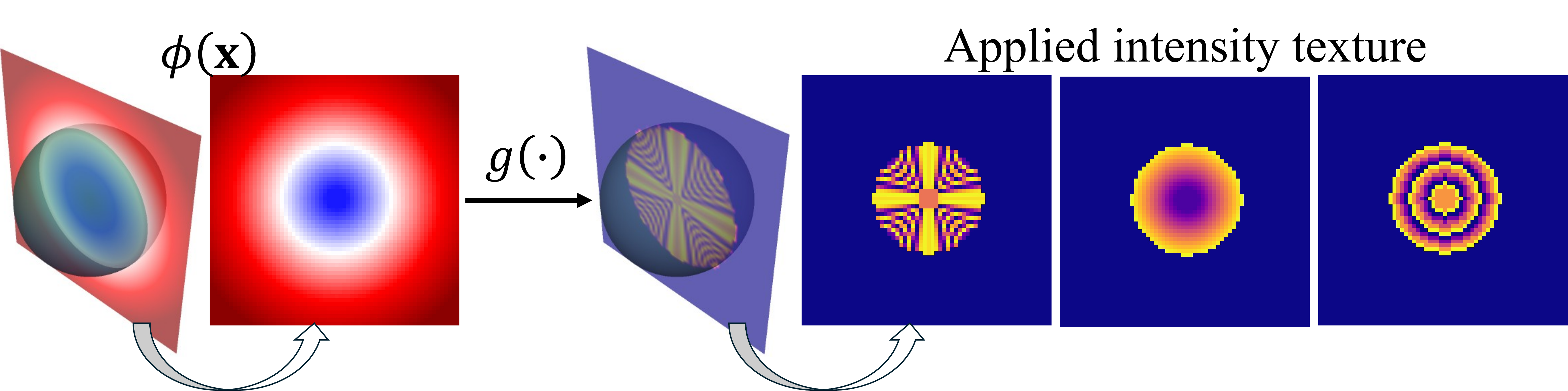}
\caption{Distance-to-intensity mapping on an SDF.
Left: SDF $\phi(\mathbf{x})$.
Mapper functions $g$ convert signed distance values into diverse intensity textures.}
\label{fig:mapper_comparison}
\end{minipage}

\end{figure}

Each displacement function $\Delta_j(\mathbf{x};\boldsymbol{\theta}_j)$ defines a continuous scalar field over $\mathbb{R}^3$, where $\boldsymbol{\theta}_j$ denotes parameters controlling geometric attributes such as frequency, amplitude, orientation, phase shift, and sharpness. As shown in Fig. \ref{fig:displacement_comparison}, displacement operates as an additive perturbation to the base signed distance field. The displaced field becomes $\phi_c(\mathbf{x}) + \Delta_j(\mathbf{x})$. The explicit functional forms of the six displacement families are provided in the Supplementary Material.

The six families include smooth sine-sum noise (Pseudo-Perlin), folded ridge-enhancing noise (Turbulence), inverted absolute-value modulation (Ridge), axis-aligned bump functions (Sharpmax), rotated stripe-like perturbations (Twisted-axis), and periodic ramp functions (Sawtooth). In practice, each of the six families is instantiated with experimentally determined parameter settings to produce visually distinct textures, yielding a total of 10 displacement variants. 

\noindent
{\bf MF Library.}
In addition to geometric variations, intensity patterns also provide important cues in medical images. To simulate diverse appearance patterns, we construct a mapper-function library that transforms signed distance values into voxel intensities.

The mapper-function library is defined as
\begin{equation}
\mathcal{F}_m
=
\{g_1, g_2, g_3, g_4, g_5, g_6\}.
\end{equation}

Each mapper $g_m(d;\boldsymbol{\psi}_m)$ maps a signed distance value $d \in \mathbb{R}$ to an intensity value, where the parameter vector $\boldsymbol{\psi}_m$ controls attributes such as decay rate, band width, frequency, and amplitude (Fig. \ref{fig:mapper_comparison}). The explicit formulations of the mapper families are provided in the Supplementary Material.

The six families include inverse-cube mapping (sharp intensity peaks near the surface), exponential decay (smooth attenuation from the boundary), linear mapping (constant intensity gradient), floor-based quantization (discrete intensity bands), modular mapping (repeating layered patterns), and sinusoidal mapping (periodic oscillations). Similarly to the DF library, each mapper family is instantiated with specific parameter configurations, producing a total of 10 mapper variants that generate visually diverse intensity patterns. 

Together, the SDF, DF, and MF libraries define the intensity generation process for a primitive object. Given a base SDF $\phi_c$, a displacement function $\Delta_j$, and a mapper function $g_m$, the resulting intensity field is expressed as
\begin{equation}
\mathbf{I}(\mathbf{x}) =
g_m\!\left(\phi_c(\mathbf{x}) + \Delta_j(\mathbf{x})\right).
\end{equation}

The complete primitive generation procedure, including spatial transformations and the composition of multiple primitives, is described in the following subsection.

\begin{algorithm}[t]
\caption{Synthetic Volume Generation via Primitive Composition}
\label{alg:data_generation}
\begin{algorithmic}[1]
\REQUIRE Libraries $\Phi, \mathcal{F}_d, \mathcal{F}_m$; Sample size $N$
\ENSURE Dataset $\mathcal{D} = \{(\mathbf{I}_i, \mathbf{Y}_i)\}_{i=1}^{N}$

\FOR{$i=1$ \TO $N$}
    \STATE Initialize $\mathbf{I}_i \leftarrow \mathbf{0}, \mathbf{Y}_i \leftarrow \mathbf{0}$ on grid $\mathcal{X}$
    \STATE $K \leftarrow$ random integer value to set number of primitive objects
    
    \FOR{$k = 1$ \TO $K$}
        \STATE Select properties $(y_k, \phi_{y_k}, \Delta_{j_k}, g_{m_k})$ and transforms $(\mathbf{R}_k, \mathbf{S}_k, \mathbf{t}_k)$ randomly
        \STATE Compute $\mathbf{x}' = (\mathbf{R}_k\mathbf{S}_k)^{-1}(\mathbf{x} - \mathbf{t}_k)$ for all $\mathbf{x} \in \mathcal{X}$
                \STATE $\mathbf{I^{(k)}}(\mathbf{x}) \leftarrow g_{m_k}(\phi_{y_k}(\mathbf{x}') + \Delta_{j_k}(\mathbf{x}'))$
        \STATE $\mathbf{M^{(k)}}(\mathbf{x}) \leftarrow \mathbb{I}[\phi_{y_k}(\mathbf{x}') + \Delta_{j_k}(\mathbf{x}') \le 0]$
        \STATE $v_k \leftarrow \text{sum}(\mathbf{M^{(k)}})$
    \ENDFOR

    \STATE $\mathbf{I}_i \leftarrow \text{Clip}(\sum_k \mathbf{I^{(k)}})$
    
    \STATE Sort indices $\{1, \dots, K\}$ by volume $v_k$ descending: $\pi(1), \dots, \pi(K)$
    \FOR{$l = 1$ \TO $K$}
        \STATE $\mathbf{Y}_i(\mathbf{x}) \leftarrow y_{\pi(l)}$ where $\mathbf{M^{(\pi(l))}}(\mathbf{x}) = 1$ \COMMENT{Smallest object wins}
    \ENDFOR
\ENDFOR
\RETURN $\mathcal{D}$
\end{algorithmic}
\end{algorithm}


\subsection{Synthetic Volume Generation via Primitive Composition}
\label{sec:data_gene}

Each synthetic volume $(\mathbf{I}_i, \mathbf{Y}_i)$ is constructed by compositing $K$ primitive objects.
For each primitive $k$, a base SDF $\phi_{y_k} \in \Phi$ is randomly selected,
where $y_k \in \{1,\dots,C\}$ denotes the class label.
In addition, a displacement function $\Delta_{j_k} \in \mathcal{F}_d$
and a mapper function $g_{m_k} \in \mathcal{F}_m$
are independently sampled from the DF and MF libraries, respectively.

Let $\mathcal{X} \subset \mathbb{R}^3$ denote the voxel grid and
$\mathbf{x} \in \mathcal{X}$ a spatial location.
To introduce geometric variability, we apply rotation $\mathbf{R}$,
shear $\mathbf{S}$, and translation $\mathbf{t}$.
The SDF is evaluated in canonical coordinates $\mathbf{x}' = (\mathbf{R}\mathbf{S})^{-1}(\mathbf{x}-\mathbf{t})$.

Surface texture and intensity patterns are jointly generated by applying
a displacement function and a mapper function to the base SDF:
\begin{equation}
\mathbf{I^{(k)}}(\mathbf{x})
=
g_{m_k}\!\left(
\phi_{y_k}(\mathbf{x}') + \Delta_{j_k}(\mathbf{x}')
\right).
\end{equation}
The corresponding binary mask is defined as
\begin{equation}
\mathbf{M^{(k)}}(\mathbf{x})
=
\mathbb{I}\!\left[ \left( \phi_{y_k}(\mathbf{x}') + \Delta_{j_k}(\mathbf{x}') \right)\le 0\right],
\end{equation}
where $\mathbb{I}[\cdot]$ denotes the indicator function, which returns 1 if the condition is satisfied and 0 otherwise.

\noindent
{\bf Merging Primitives.}
Given $\{(\mathbf{I}^{(k)}, \mathbf{M}^{(k)}, y_k)\}_{k=1}^{K}$,
we compose the intensities by summation:
\begin{equation}
\mathbf{I}_i
=
\sum_{k=1}^{K}
\mathbf{I}^{(k)} .
\end{equation}

In practice, the resulting intensities are clipped to the valid range.

Following \cite{tadokoro2024primitive}, labels are assigned by prioritizing primitives with smaller mask areas. In overlapping regions, the class label of the primitive with the smaller mask is selected.

Let $v_k = \sum_{\mathbf{x}} \mathbf{M}^{(k)}(\mathbf{x})$ denote the mask volume of the $k$-th primitive, and let $\pi$ be a permutation such that
$v_{\pi(1)} \ge \cdots \ge v_{\pi(K)}$.
The label map is defined as
\begin{equation}
\mathbf{Y}_i(\mathbf{x})
=
\begin{cases}
y_{\pi \left(
\max \{\, l \mid \mathbf{M}^{(\pi(l))}(\mathbf{x}) = 1 \,\}
\right)}
& \text{if } \exists\, l \text{ with } \mathbf{M}^{(\pi(l))}(\mathbf{x}) = 1, \\
0 & \text{otherwise}.
\end{cases}
\end{equation}

\noindent
{\bf Dataset Construction and Pre-training.}
The above procedure generates one synthetic sample $(\mathbf{I}_i,\mathbf{Y}_i)$.
By repeating this process $N$ times with independently sampled primitives,
transformations, and texture functions, we construct a synthetic dataset
$\mathcal{D}=\{(\mathbf{I}_i,\mathbf{Y}_i)\}_{i=1}^{N}$.

This dataset provides volumetric images and corresponding voxel-wise labels,
enabling supervised training of a 3D segmentation model.
We therefore use $\mathcal{D}$ to pre-train the network,
and subsequently fine-tune the model on real downstream datasets.

\section{Experiment}
\label{sec:experiment}

\subsection{Experimental Setup}
\label{sec:exp_setup}

\noindent
{\bf Datasets.}
Following~\cite{Isensee2024-vg}, we evaluated FDIF on three segmentation datasets: AMOS22~\cite{ji2022amos}, ACDC~\cite{bernard2018deep}, and KiTS19~\cite{heller2023kits21}. AMOS22 contains CT and MRI scans annotated for 15 abdominal organs, ACDC is a cine-MRI dataset for cardiac segmentation with three structures, and KiTS19 provides CT scans for kidney and renal tumor segmentation. Following~\cite{wald2025openmind}, all datasets were split into 50\%/50\% train/validation sets.

For classification, we used MRNet~\cite{bien2018deep} and MedMNIST~\cite{medmnistv2}. MRNet is a knee MRI dataset for multi-label abnormality classification using T1-, T2-, and PD-weighted sequences. From MedMNIST, we used three 3D datasets at $64\times64\times64$: OrganMNIST3D~\cite{organmnist2}, NoduleMNIST3D~\cite{nodulemnist3d}, and FractureMNIST3D~\cite{fracturemnist3d}. Default train/validation splits were used for classification experiments.

\noindent
{\bf Architectures.}
To evaluate robustness across model designs, we used three segmentation architectures: SwinUNETR~\cite{hatamizadeh2021swin}, nnUNet ResEnc-L (Residual Encoder U-Net)~\cite{Isensee2024-vg}, and nnUNet Primus-M~\cite{wald2025primus}. For SwinUNETR, we followed~\cite{tang2022selfsupervisedpretrainingswintransformers,tadokoro2024primitive} with feature size 48 and a patch size of $96\times96\times96$. For nnUNet ResEnc-L and Primus-M, we adopted the configuration of~\cite{Wald2024-ds}, using a patch size of $160\times160\times160$, resampling inputs to $1\,\text{mm}\times1\,\text{mm}\times1\,\text{mm}$, and applying z-score normalization.

For classification, we used nnUNet ResEnc-L with the same configuration as in segmentation, except that the input patch size was adjusted to cover the full volume of each dataset.

\noindent
{\bf Implementation Details.}
All FDIF data were generated at a resolution of $96\times96\times96$. For segmentation, 5{,}000 samples were generated per FDIF configuration, each containing 20 objects. Unless otherwise stated, the default FDIF setting used 109 global shapes with 10 mapper and 10 displacement variants. As a baseline, PrimGeoSeg data were generated under the same conditions with a fixed class count of 32.

For SwinUNETR, pre-training used a batch size of 4 with gradient accumulation of 2, optimized with AdamW (learning rate $1\times10^{-4}$, weight decay $1\times10^{-5}$) and a WarmupCosineAnnealing schedule for 200{,}000 iterations. During fine-tuning, the output layer was replaced to match downstream classes, and the model was trained end-to-end for 15{,}000 iterations with the same optimizer.

For nnUNet ResEnc-L and Primus-M, we used the public nnUNet framework. Pre-training used a batch size of 8 with SGD (initial learning rate $1\times10^{-2}$) and a polynomial schedule for 500 epochs (250 iterations per epoch). During fine-tuning, the output layer was replaced and the remaining weights were initialized from pre-training. Training used AdamW for 150 epochs with a batch size of 2 and an initial learning rate of $1\times10^{-3}$, with a Sawtooth scheduler for ResEnc-L and a Warmup scheduler for Primus-M.

For classification, we used the nnUNet classification framework. Pre-training and fine-tuning both used AdamW with cosine annealing (initial learning rate $1\times10^{-2}$). The FDIF classification dataset contained 50 samples per class (109 global-shape classes). Pre-training ran for 100 epochs, followed by 200 epochs of fine-tuning for each downstream dataset.

\subsection{Segmentation Results}
\label{sec:exp_segmentation}

\begin{table}[!t]
\centering
\caption{Comparison of segmentation performance in terms of Dice score across different architectures and pre-training methods on three benchmark datasets. Best results are highlighted in \textbf{bold} and second-best results are \underline{underlined}.}
\label{tab:comparison}

\begin{tabular}{llccccc}
\toprule
Architecture & Pre-training & Type & AMOS & ACDC & KiTS & Avg \\
\midrule

\multirow{4}{*}{SwinUNETR}
 & Scratch & - & 80.98 & 81.91 & 74.55 & 79.15 \\
\cdashline{2-7}
 & PrimGeoSeg\cite{tadokoro2024primitive} & \multirow{3}{*}{FDSL} & 81.89 & \textbf{85.02} & 75.80 & 80.90 \\
 & Ours (FDIF, Disp+Map) &  & \underline{82.28} & 83.82 & \underline{77.20} & \underline{81.10} \\
 & Ours (FDIF, Map only) &  & \textbf{82.70} & \underline{84.66} & \textbf{78.31} & \textbf{81.89} \\
\midrule

\multirow{7}{*}{nnUNet ResEnc-L}
 & Scratch & - & 85.97 & 91.98 & 83.84 & 87.26 \\
\cdashline{2-7}
 & S3D\cite{Wald2024-ds, wald2025openmind} & \multirow{3}{*}{SSL} & 86.16 & 92.01 & 86.01 & 88.06 \\
 & MG\cite{zhou2021models, wald2025openmind} &  & 86.35 & 91.74 & \textbf{86.17} & 88.09 \\

 & MAE\cite{he2022masked, wald2025openmind} &  & 86.78 & 91.98 & \underline{86.12} & 88.30 \\
\cdashline{2-7}
 & PrimGeoSeg\cite{tadokoro2024primitive} & \multirow{3}{*}{FDSL} & 87.57 & 92.26 & 85.74 & 88.52 \\
 & Ours (FDIF, Disp+Map) &  & \underline{87.66} & \textbf{92.43} & 86.02 & \textbf{88.70} \\
 & Ours (FDIF, Map only) &  & \textbf{88.04} & \underline{92.36} & 85.57 & \underline{88.65} \\
\midrule

\multirow{7}{*}{nnUNet Primus-M}
 & Scratch & - & 84.00 & 91.61 & 76.92 & 84.18 \\
\cdashline{2-7}
 & VF\cite{wang2023mis, wald2025openmind} & \multirow{3}{*}{SSL} & 84.95 & 91.41 & 86.17 & 87.51 \\
 & SimMIM\cite{chen2023masked, wald2025openmind} &  & 86.57 & 91.98 & 85.92 & 88.16 \\
 & MAE\cite{he2022masked, wald2025openmind} & & \underline{87.16} & \underline{92.16} & \textbf{86.74} & \textbf{88.69} \\
\cdashline{2-7}
 & PrimGeoSeg\cite{tadokoro2024primitive} & \multirow{3}{*}{FDSL} & 83.70 & \underline{92.16} & 82.34 & 86.07 \\
 & Ours (FDIF, Disp+Map) &  & 86.91 & \underline{92.16} & 86.37 & 88.48 \\
 & Ours (FDIF, Map only) &  & \textbf{87.26} & \textbf{92.24} & \underline{86.45} & \underline{88.65} \\
\bottomrule
\end{tabular}
\end{table}

We compared our method with three training strategies: training from scratch (without pre-training), formula-driven supervised learning (FDSL), and self-supervised learning (SSL).
As an FDSL baseline, we included PrimGeoSeg~\cite{tadokoro2024primitive}, which pre-trains segmentation networks using synthetic geometric primitives. For SSL, we selected representative methods based on the benchmark results reported in~\cite{wald2025openmind}: MAE~\cite{he2022masked}, which reconstructs masked image patches; Models Genesis (MG)~\cite{zhou2021models}, which learns representations via restoration-based pretext tasks for 3D medical images; S3D~\cite{Wald2024-ds}, which introduces sparse masked reconstruction for CNNs; SimMIM~\cite{chen2023masked}, which predicts raw pixel values of masked regions; and VolumeFusion (VF)~\cite{wang2023mis}, which constructs segmentation-aware pretext tasks from unlabeled volumes.

Table~\ref{tab:comparison} reports the Dice scores on AMOS, ACDC, and KiTS using three architectures: SwinUNETR, nnUNet ResEnc-L, and nnUNet Primus-M. Overall, FDIF achieved competitive or superior performance across all architectures and datasets\footnote{Results of the self-supervised methods are taken from the benchmark results reported in~\cite{wald2025openmind}.}.

Notably, FDIF improved over PrimGeoSeg, the current state-of-the-art method in formula-driven supervised learning (FDSL), by 2.58 Dice points on nnUNet Primus-M (88.65 vs. 86.07), demonstrating substantial progress within the FDSL paradigm.
Furthermore, on nnUNet ResEnc-L, FDIF surpassed MAE by 0.41 Dice points in terms of average performance. This result indicates that our approach can match or even exceed self-supervised models pre-trained on large-scale real data.

\begin{figure}[!t]
\centering
\includegraphics[width=\textwidth]{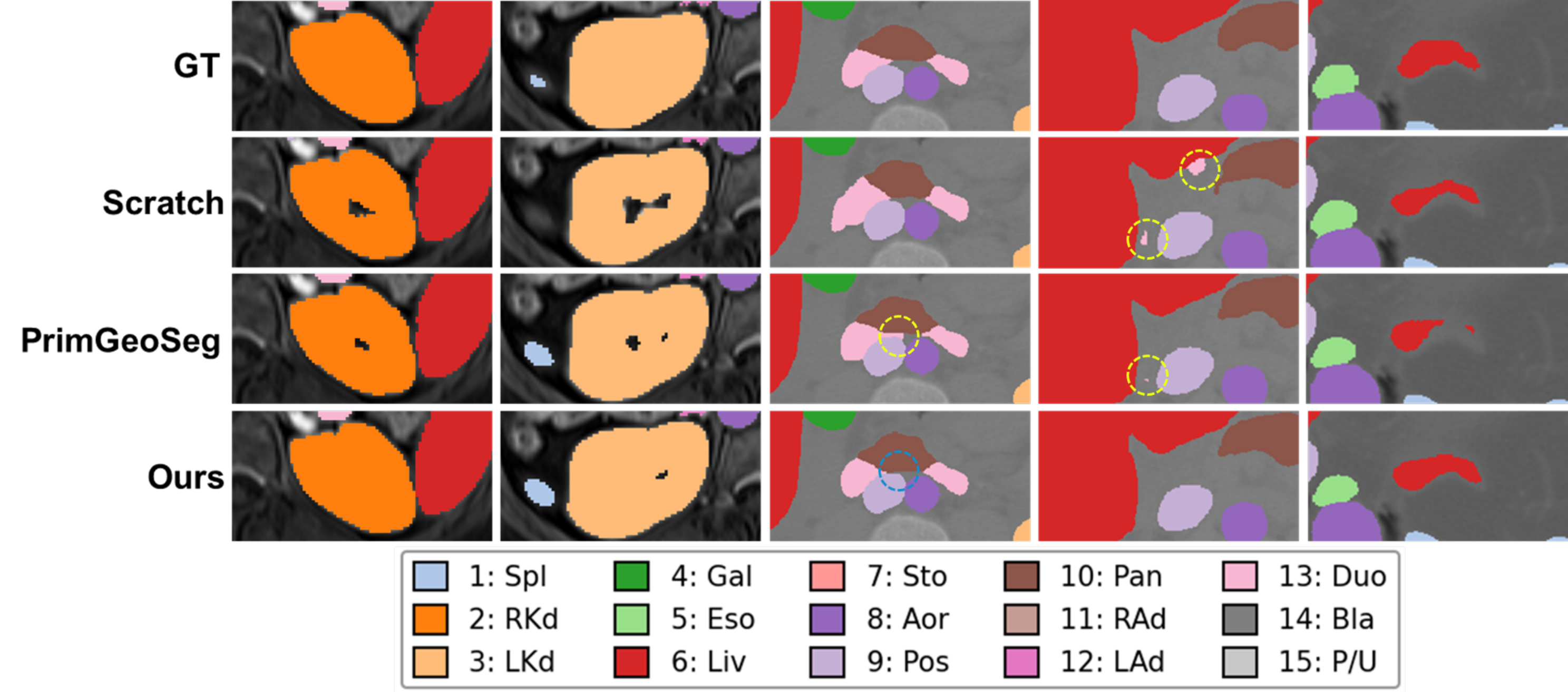}
\caption{Qualitative comparison of segmentation results. Models pre-trained with FDIF demonstrate improved segmentation accuracy compared to baseline methods.}
\label{fig:experiment_results}
\end{figure}

Fig.~\ref{fig:experiment_results} shows qualitative segmentation results. Compared with PrimGeoSeg, FDIF reduces both over-detection and miss-detection. For example, PrimGeoSeg over-detects the duodenum (third column), whereas FDIF suppresses this false positive. In addition, PrimGeoSeg misses part of the liver (fifth column), while FDIF correctly segments it. These results indicate that FDIF pre-training learns richer shape and texture representations, leading to more accurate boundary delineation.

For the transformer-based SwinUNETR, the Map-only variant of our method achieved the highest average score of 81.89, outperforming PrimGeoSeg (80.90) and the model trained from scratch (79.15). For the CNN-based nnUNet ResEnc-L, the Disp+Map configuration achieved the best average score of 88.70, surpassing existing SSL approaches including MAE (88.30), Models Genesis (88.09), and S3D (88.06). Compared with training from scratch, FDIF consistently improved segmentation accuracy across all evaluated architectures, indicating that the proposed pre-training strategy provides effective initialization and enhances feature representations for downstream segmentation tasks.

Finally, although MAE achieved the best average score for nnUNet Primus-M (88.69), the difference from our method (88.65) was negligible. Overall, the results indicate that FDIF pre-training achieves strong and stable performance across diverse architectures and datasets. Importantly, these results were obtained without requiring any real images during pre-training, highlighting the effectiveness of FDIF as a practical and scalable alternative to conventional self-supervised pre-training strategies.

\subsection{Ablation Study}
\label{sec:exp_ablation}

Table~\ref{tab:ablation} analyzes the impact of the global-shape count, displacement augmentation, and mapper functions on the AMOS dataset.
In configurations without multiple mappers (10 global shapes, 109 global shapes, and 109,gs+10,Disp), the Inverse Cube mapper was applied uniformly to all samples.

All FDIF configurations outperformed the PrimGeoSeg baseline, confirming the overall effectiveness of FDIF. Increasing the number of global shapes from 10 to 109 provided only a marginal improvement of 0.01 Dice points. In contrast, introducing texture diversity through displacement or mapper augmentation yielded larger gains of 0.06 and 0.08 points, respectively. These results indicate that texture diversity contributes more to representation learning than simply increasing the number of global shapes.

Interestingly, the combined Disp+Map configuration yielded the lowest score among FDIF variants on AMOS. However, Table~\ref{tab:comparison} shows that this trend does not generalize across datasets. The lower AMOS score was mainly due to the Bladder class, which achieved its best result under the 109 global-shape configuration without texture augmentation. This suggests that accurate recognition of this organ benefits more from global shape understanding than from additional texture variations. Consequently, allocating model capacity to both displacement and mapper textures in the 109,gs+10,Disp+10,Map configuration may lead to a relatively weaker representation of global shape. Overall, the optimal FDIF configuration depends on the characteristics of the downstream task.

\begin{table}[!t]
\centering
\caption{Ablation study on the AMOS dataset evaluating the effects of global-shape count, displacement, and mapper functions. Results are reported as Dice scores (\%). Organ abbreviations: Spl=Spleen, RKd=Right Kidney, LKd=Left Kidney, Gal=Gallbladder, Eso=Esophagus, Liv=Liver, Sto=Stomach, Aor=Aorta, Pos=Postcava, Pan=Pancreas, RAd=Right Adrenal, LAd=Left Adrenal, Duo=Duodenum, Bla=Bladder, P/U=Prostate/Uterus.}
\label{tab:ablation}
\resizebox{\textwidth}{!}{%
\begin{tabular}{l|c|ccccccccccccccc}
\toprule
Pre-training & Avg & Spl & RKd & LKd & Gal & Eso & Liv & Sto & Aor & Pos & Pan & RAd & LAd & Duo & Bla & P/U \\
\midrule
10 global-shape & 87.95 & 96.84 & 96.00 & \textbf{95.58} & 83.32 & 82.59 & 97.65 & 91.17 & \textbf{94.91} & 89.87 & 87.15 & 75.90 & 77.27 & 81.09 & 86.90 & 82.95 \\
109 global-shape & 87.96 & \textbf{96.89} & \textbf{96.20} & 95.03 & 84.36 & \textbf{83.03} & 97.67 & 90.80 & 94.68 & 90.13 & 87.09 & \textbf{76.11} & \textbf{77.38} & 80.89 & \textbf{87.11} & 82.01 \\
109 gs+10 Disp & 88.02 & 96.77 & 95.96 & 95.49 & \textbf{84.66} & 82.79 & 97.68 & \textbf{91.10} & 94.87 & \textbf{90.24} & 87.34 & 75.50 & 77.62 & 81.51 & 86.77 & 82.00 \\
109 gs+10 Map & \textbf{88.04} & 96.86 & 96.14 & 95.47 & 84.19 & 83.00 & \textbf{97.69} & 90.93 & 94.87 & 90.21 & \textbf{87.49} & 76.07 & 77.47 & 81.53 & 86.24 & 82.45 \\
109 gs+10 Disp+10 Map & 87.66 & 96.78 & 96.06 & 95.55 & 84.11 & 82.53 & \textbf{97.69} & 90.58 & 94.82 & \textbf{90.24} & 87.46 & 76.06 & 77.41 & \textbf{81.59} & 81.15 & 82.89 \\
\midrule
PrimGeoSeg (32 gs) & 87.57 & 96.80 & 95.76 & 95.44 & 83.20 & 81.96 & 97.56 & 90.37 & 94.78 & 89.96 & 87.04 & 75.85 & 76.89 & 80.36 & 85.62 & 82.01 \\
Scratch & 85.97 & 96.42 & 95.20 & 95.00 & 79.75 & 80.68 & 97.26 & 88.84 & 94.42 & 89.70 & 84.87 & 74.70 & 74.88 & 78.62 & 81.30 & 77.96 \\
\bottomrule
\end{tabular}%
}
\end{table}

\begin{table}[!t]
\centering
\caption{Ablation study on global-shape construction methods: Comparison between extrusion-only, revolution/hollowing-only, and combined approaches on AMOS dataset.}
\label{tab:ablation_shape}
\resizebox{\textwidth}{!}{%
\begin{tabular}{l|c|ccccccccccccccc}
\toprule
Pre-training & Avg & Spl & RKd & LKd & Gal & Eso & Liv & Sto & Aor & Pos & Pan & RAd & LAd & Duo & Bla & P/U \\
\midrule
10 gs (Extrusion only) & 87.80 & \textbf{96.93} & 95.82 & 94.90 & 83.31 & 82.37 & 97.62 & 90.49 & 94.86 & 89.79 & 87.01 & 75.88 & 77.14 & \textbf{81.13} & 86.34 & \textbf{83.36} \\
10 gs (Revolution \& Hollowing only) & 87.78 & 96.85 & \textbf{96.20} & 95.42 & \textbf{83.92} & 82.24 & 97.62 & 90.93 & 94.76 & 89.86 & \textbf{87.20} & \textbf{76.25} & 76.99 & 81.03 & 85.40 & 82.07 \\
10 gs (Combined) & \textbf{87.95} & 96.84 & 96.00 & \textbf{95.58} & 83.32 & \textbf{82.59} & \textbf{97.65} & \textbf{91.17} & \textbf{94.91} & \textbf{89.87} & 87.15 & 75.90 & \textbf{77.27} & 81.09 & \textbf{86.90} & 82.95 \\
\bottomrule
\end{tabular}%
}
\end{table}

Table~\ref{tab:ablation_shape} further shows that combining diverse shape construction types is more beneficial than simply increasing the number of shapes.
We compared three settings without displacement or mapper augmentation: Extrusion only (similar to PrimGeoSeg), Revolution \& Hollowing only (shapes not representable by extrusion), and a Combined configuration where 10 shapes were randomly sampled from both construction types.
The Combined configuration outperformed both single-type settings by 0.15 Dice points, which is larger than the gain obtained by increasing the number of global shapes from 10 to 109.

\subsection{Extension to 3D Classification}
\label{sec:exp_classification}

The use of FDIF-generated synthetic data for model pre-training is not limited to segmentation tasks.
Any task that relies on 3D shape and texture information may potentially benefit from FDIF pre-training.
To investigate this possibility, we extended FDIF to the 3D classification setting.

For this experiment, we adapted the FDIF generation pipeline by placing a single object at the center of each volume and assigning its global-shape class ID as the sample label.
Random rotations and shear transformations were applied to diversify object poses, while translation was disabled so that the object remained centered.
The label of each sample therefore corresponded to the global-shape class of the contained object.
As in the segmentation setting, displacement and mapper functions were used as augmentations, allowing objects of the same class to exhibit diverse local geometric and intensity textures.
This configuration encourages the model to learn global shape contours from intensity and texture patterns, while also recognizing the boundary between the object and the background.
As a result, the model learns to predict the shape category directly from its 3D appearance.
As an SSL baseline, we used the publicly available SwinUNETR pre-trained weights released by~\cite{wald2025openmind}, which were fine-tuned on each downstream dataset under the same protocol used for FDIF.

As shown in Table~\ref{tab:classification}, FDIF achieved the highest average accuracy across the four classification benchmarks, outperforming both training from scratch and SSL pre-training with SwinUNETR~\cite{tang2022selfsupervisedpretrainingswintransformers}.
An exception was the MedMNIST Nodule dataset, where FDIF yielded the lowest score among the pre-training strategies while training from scratch achieved the highest accuracy, suggesting that this task may not benefit significantly from pre-training.
Nevertheless, FDIF improved the average accuracy by 1.32 points compared with SwinUNETR SSL.
Although our evaluation is limited in scope, these results suggest that implicit-function-based pre-training may generalize beyond segmentation and potentially benefit a broader range of 3D recognition tasks.

\begin{table}[!t]
\centering
\caption{Transfer learning to 3D classification tasks. Balanced Accuracy (\%) is reported for each dataset. Models are pre-trained with different strategies and fine-tuned on four classification benchmarks.}
\label{tab:classification}
\begin{tabular}{lccccc}
\toprule
Pre-training & MRNet & \makecell{MedMNIST\\Organ} & \makecell{MedMNIST\\Nodule} & \makecell{MedMNIST\\Fracture} & Avg \\
\midrule
Ours (FDIF) & \textbf{67.45} & \textbf{96.52} & 79.72 & \textbf{69.21} & \textbf{78.23} \\
SwinUNETR (SSL)\cite{tang2022selfsupervisedpretrainingswintransformers} & 67.36 & 92.05 & 80.48 & 67.75 & 76.91 \\
Scratch & 62.19 & 85.32 & \textbf{85.12} & 56.22 & 72.21 \\
\bottomrule
\end{tabular}
\end{table}

\section{Conclusion}
We presented FDIF, a pre-training framework based on fractal procedural implicit functions for learning 3D representations from fully synthetic data. By combining global shape primitives with displacement and distance-to-intensity mapper functions, FDIF generates diverse volumetric samples without relying on real images.
Experiments on multiple medical image segmentation benchmarks show that FDIF provides effective initialization for 3D models and achieves competitive performance compared with existing pre-training approaches. Additional studies indicate that increasing geometric and texture diversity plays an important role in representation learning. We further demonstrate that FDIF can also benefit 3D classification tasks, suggesting that the learned representations transfer beyond segmentation.
These results highlight the potential of procedural implicit-function generation as a scalable alternative to data-driven pre-training. We hope FDIF encourages further research on synthetic data generation and implicit representations for general 3D learning.

\section*{Acknowledgements}
Computational resources of AI Bridging Cloud Infrastructure (ABCI) and ABCI-Q provided by National Institute of Advanced Industrial Science and Technology (AIST) were used. We would like to thank Ryu Tadokoro and Kazuma Kobayashi for their helpful research discussion.

%

\bibliographystyle{splncs04}
\bibliography{main2}
\end{document}


\title{Supplementary Materials\\FDIF: Formula-Driven supervised Learning \\ with Implicit Functions \\for 3D Medical Image Segmentation}

\author{Yukinori Yamamoto\inst{1, 3}\thanks{\email{yamanokoai@gmail.com}} \and Kazuya Nishimura\inst{2} \and Tsukasa Fukusato\inst{1} \and\\ Hirokazu Nosato\inst{3} \and Tetsuya Ogata\inst{1, 3} \and Hirokatsu Kataoka\inst{3, 4}}

\authorrunning{Y.~Yamamoto et al.}

\institute{Waseda University \and The University of Osaka \and National Institute of Advanced Industrial Science and Technology \and University of Oxford}

\maketitle

\appendix

\section{SDF Library Details}
\label{suppl:sdf}

We provide closed-form signed distance function (SDF) definitions for the shape library used in FPIF.
All SDFs assign a negative value to interior points, zero on the boundary, and a positive value to exterior points.
Our SDF implementations are based on the distance function catalogue by Qu{\'i}lez~\cite{quilez2015distancefunctions,10.1145/2659467.2659474}.
The library contains $\mathcal{C}=109$ classes: three native 3D primitives (sphere, octahedron, cone) and 106 classes generated by applying extrusion, revolution, or hollowing to 2D base shapes.

\subsection{Native 3D Primitives}
\label{suppl:sdf_primitives}

\noindent{\bf Sphere.}
A sphere of radius $r$ centered at the origin:
\begin{equation}
\phi_{\text{sphere}}(\mathbf{p}) = \|\mathbf{p}\| - r,
\end{equation}
where $\mathbf{p}=(x,y,z)^\top$.

\noindent{\bf Octahedron.}
A regular octahedron with ``radius'' (half-diagonal) $s$:
\begin{equation}
\phi_{\text{oct}}(\mathbf{p}) = \frac{|x|+|y|+|z|-s}{\sqrt{3}}.
\end{equation}
The factor $1/\sqrt{3}$ normalises the gradient to unit length along the face normals.

\noindent{\bf Cone.}
A cone with half-angle $\theta$ and height $h$, with apex at $(0,h,0)$ and base disk at $y=0$.
Let $\rho=\sqrt{x^2+z^2}$ be the radial distance from the $y$-axis and
$\mathbf{w}=(\rho,\,y)^\top$ the 2D representative point.
Define the 2D slant-edge direction
\begin{equation}
\mathbf{q} = \begin{pmatrix}h\tan\theta \\ -h\end{pmatrix}.
\end{equation}
Then compute clamped-projection residuals onto the slant edge and the base cap:
\begin{align}
\mathbf{a} &= \mathbf{w} - \mathbf{q}\,\mathrm{clamp}\!\left(\frac{\mathbf{w}\cdot\mathbf{q}}{\|\mathbf{q}\|^2},\;0,\;1\right),\\
\mathbf{b} &= \mathbf{w} - \begin{pmatrix}\mathrm{clamp}\!\left(\rho/q_1,\,0,\,1\right)q_1\\ q_2\end{pmatrix},
\end{align}
where $q_1=h\tan\theta$ and $q_2=-h$.  The signed distance is
\begin{equation}
\phi_{\text{cone}}(\mathbf{p})
= \sqrt{\min\!\left(\|\mathbf{a}\|^2,\,\|\mathbf{b}\|^2\right)}
\cdot\mathrm{sign}(s),
\end{equation}
where the sign discriminant (using $k=\mathrm{sign}(q_2)=-1$ and writing $w_1=\rho$, $w_2=y$ for the components of $\mathbf{w}$) is
\begin{equation}
s = \max\!\bigl(k(w_1 q_2 - w_2 q_1),\;k(w_2 - q_2)\bigr).
\end{equation}

\subsection{2D Base Shape SDFs}
\label{suppl:sdf_2d}

The remaining 106 shape classes are derived from 2D SDFs via the 3D construction methods described in Sec.~\ref{suppl:sdf_3d}.
We use 11 distinct 2D base shapes: 7 irregular polygon classes and 4 star-shape classes.

\noindent{\bf Polygons.}
An $n$-vertex irregular polygon ($n\in\{3,\ldots,9\}$) is generated by sampling one vertex per angular sector.
Sector $i$ covers $[2\pi i/n,\;2\pi(i+1)/n)$; its vertex is placed at a random angle within the sector and a random radius $r_i\in[r_{\min},r_{\max}]$.
This guarantees a non-self-intersecting contour with controlled randomness.
For a 2D query point $\mathbf{u}=(u,v)$, the SDF is
\begin{equation}
\phi_{\text{poly}}(\mathbf{u})
= \sigma(\mathbf{u})\cdot\min_{e}\,d(\mathbf{u},e),
\end{equation}
where the minimum is taken over all polygon edges $e$, $d(\mathbf{u},e)$ is the Euclidean distance from $\mathbf{u}$ to edge $e$, and $\sigma(\mathbf{u})\in\{-1,+1\}$ is the sign determined by a winding-number test: $\sigma=-1$ if $\mathbf{u}$ is inside the polygon, $\sigma=+1$ otherwise.

\noindent{\bf Stars.}
Star shapes extend the 2D vocabulary to higher vertex counts ($n\in\{5,\ldots,8\}$ arms) while preserving sharp angular variations---a property that regular $n$-gons lose for $n\ge 10$ as they approach a circle.
Each star is parameterised by the number of arms $n$, a concavity parameter $w\in[0,1]$ controlling the depth of inward cusps, and a scale $r$.
The SDF is computed by folding the input coordinates through angular ($2\pi/n$ rotational symmetry) and radial symmetry operations, projecting the folded point onto the representative sector boundary, and applying a clamped distance metric whose sign is derived from the angular position within the sector.
The 4 star classes (with $n=5,6,7,8$) together with the 7 polygon classes yield 11 distinct 2D base shapes.

\subsection{3D Shape Construction}
\label{suppl:sdf_3d}
\begin{figure}[!t]
\centering
\includegraphics[width=\textwidth]{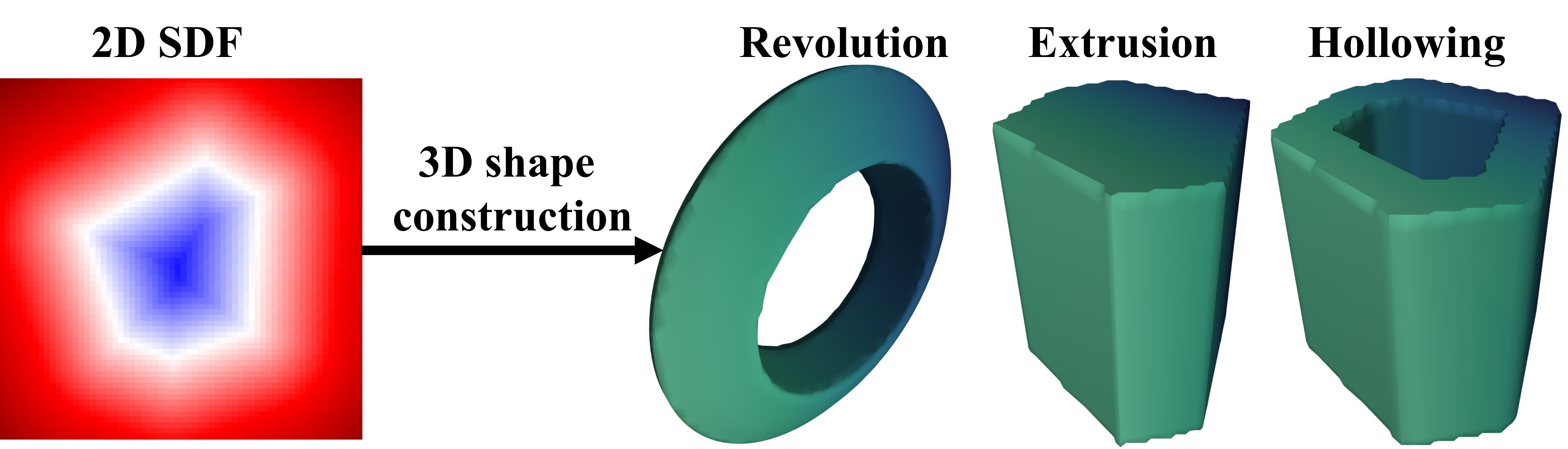}
\caption{SDF-based 3D shape construction methods. We generate 106 classes by applying extrusion, revolution, or hollowing to 2D base shapes. Each method is parameterised to produce multiple variants per base shape, yielding a rich and diverse shape library.}
\label{fig:data_generation_flow}
\end{figure}

\noindent{\bf Extrusion with $z$-dependent scaling.}
Let $f(x,y)$ be a 2D SDF (negative inside), half-height $h>0$, and $s(z)>0$ a scaling profile.
The cross-sectional distance in the $xy$-plane at elevation $z$ and the along-axis distance from the slab are
\begin{align}
\phi_\perp(x,y,z) &= s(z)\,f\!\left(\frac{x}{s(z)},\frac{y}{s(z)}\right),\\
\phi_z(x,y,z) &= |z| - h.
\end{align}
These are combined using the standard SDF exterior-only overlap (equivalent to an intersection in SDF algebra):
\begin{equation}
\phi_{\text{extr}}(x,y,z)
= \underbrace{\sqrt{\max(\phi_\perp,0)^2+\max(\phi_z,0)^2}}_{\text{exterior corner}}
+ \underbrace{\min\!\bigl(\max(\phi_\perp,\phi_z),\,0\bigr)}_{\text{interior}},
\end{equation}
which correctly handles all four quadrants of the $\phi_\perp$--$\phi_z$ plane.
Taking $s(z)=\text{const}$ yields a uniform prism; varying $s(z)$ linearly or with a smooth profile produces cones, tapered solids, or bulged forms.

\noindent{\bf Revolution.}
Given any 2D SDF $f(u,v)$ (negative inside), the surface of revolution around the $y$-axis with major radius $R\ge 0$ is obtained by the coordinate substitution
\begin{equation}
\rho = \sqrt{x^2+z^2},\quad u = \rho - R,\quad v = y,\quad
\phi_{\text{rev}}(x,y,z) = f(u,\,v).
\end{equation}
For $R>0$ this sweeps the 2D profile into a torus-like solid; for $R=0$ it yields a solid of revolution.

\noindent{\bf Hollowing (onionization).}
To create tube-like or shell-like structures from any closed SDF $\phi$, we apply repeated absolute-value folding followed by thickness subtraction.
Given a list of thickness values $\{t_j\}_{j=1}^{m}$, the hollowing operation is
\begin{equation}
\phi \leftarrow |\phi|,\qquad
\phi \leftarrow |\phi| - t_j\quad\text{for each }t_j.
\end{equation}
Each fold creates one additional concentric shell; the result is a multi-layered shell structure whose layer widths are controlled by $\{t_j\}$.

\section{Displacement Function Details}
\label{suppl:displacement}
\begin{figure}[!t]
\centering
\includegraphics[width=\textwidth]{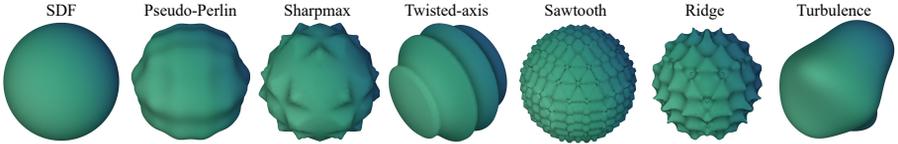}
\caption{Displacement function families. We apply a variety of procedural displacement functions to the base shapes, generating diverse surface details that challenge the generalization capabilities of SDF regression models. Each family is parameterised to produce multiple variants, resulting in a rich dataset of 3D shapes with complex surface patterns.}
\label{fig:data_generation_flow}
\end{figure}

We provide detailed mathematical formulations for all displacement functions used in FPIF. Each displacement function $\Delta:\mathbb{R}^3\to\mathbb{R}$ is applied to normalized coordinates $\tilde{\mathbf{x}}=(\tilde{x},\tilde{y},\tilde{z})\in[-1,1]^3$ and scaled by an amplitude $A>0$.

\noindent{\bf Pseudo-Perlin.}
We approximate smooth noise with a sum of $J$ sinusoidal components:
\begin{equation}
\Delta_{\text{perlin}}(\tilde{\mathbf{x}})
=
A\sum_{j=1}^{J} a_j\,
\sin(2\pi\,\mathbf{k}_j^{\top}\tilde{\mathbf{x}} + b_j),
\end{equation}
where $J$ is the number of terms, $a_j$ are decreasing amplitudes, $\mathbf{k}_j\in\mathbb{R}^3$ are frequency vectors (scaled by a base frequency $f$), and $b_j$ are phase offsets. Two variants are obtained by using different $f$ and/or $\{a_j\}$.

\noindent{\bf Sharpmax.}
Sharp, axis-aligned bumps are produced by taking the maximum of absolute stripes:
\begin{equation}
\begin{aligned}
\Delta_{\text{sharpmax}}(\tilde{\mathbf{x}})
=A\max\big(&|\sin(2\pi f\tilde{x})|,\\
&|\sin(2\pi f\tilde{y})|,\\
&|\sin(2\pi f\tilde{z})|\big).
\end{aligned}
\end{equation}

\noindent{\bf Twisted-axis.}
We generate stripes in a rotated local frame whose angle depends on another axis. For example, twisting $x$-stripes along $\tilde{z}$:
\begin{align}
\tilde{x}_\theta(\tilde{\mathbf{x}}) &= \tilde{x}\cos(\kappa \tilde{z}) - \tilde{y}\sin(\kappa \tilde{z}),\\
\Delta_{\text{twist}}(\tilde{\mathbf{x}}) &= A\left|\sin(2\pi f\,\tilde{x}_\theta)\right|,
\end{align}
where $\kappa$ controls the twist rate.

\noindent{\bf Sawtooth.}
Let $\mathrm{frac}(t)=t-\lfloor t\rfloor$ and define a sawtooth wave $\mathrm{saw}(t)=2\,\mathrm{frac}(t)-1$. Then:
\begin{equation}
\Delta_{\text{saw}}(\tilde{\mathbf{x}})=A\,\mathrm{saw}(f\tilde{x}),
\end{equation}
and similarly for $\tilde{y}$ or $\tilde{z}$. Two variants are obtained by using different $f$.

\noindent{\bf Ridge.}
Ridge-like patterns are produced by inverting the absolute magnitude of smooth noise:
\begin{equation}
\Delta_{\text{ridge}}(\tilde{\mathbf{x}})
=
A\left(1-\left|\frac{\Delta_{\text{perlin}}(\tilde{\mathbf{x}})}{A}\right|\right).
\end{equation}
Two ridge variants are obtained by using different Perlin parameter sets.

\noindent{\bf Turbulence.}
Turbulence folds the smooth components to emphasize higher-frequency ridges:
\begin{equation}
\Delta_{\text{turb}}(\tilde{\mathbf{x}})
=
A\sum_{j=1}^{J} a_j\,
\left|
\sin(2\pi\,\mathbf{k}_j^{\top}\tilde{\mathbf{x}} + b_j)
\right|.
\end{equation}

\section{Mapper Function Details}
\label{suppl:mapper}
\begin{figure}[!t]
\centering
\includegraphics[width=\textwidth]{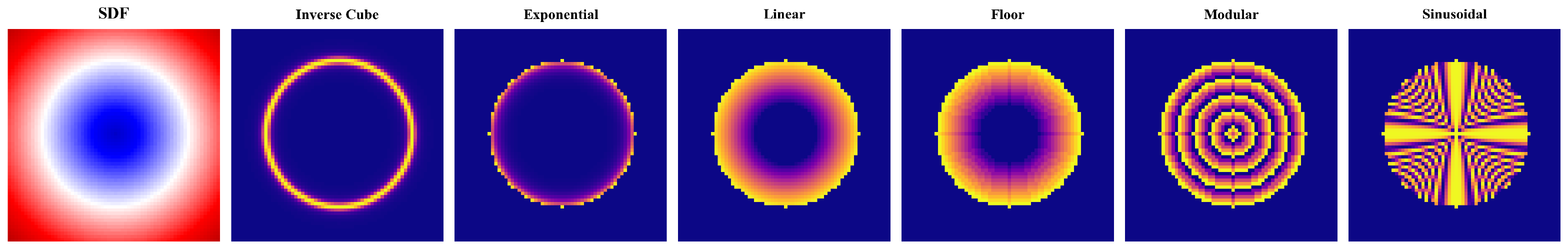}
\caption{Mapper function families. We apply a variety of mapper functions to convert signed distance values into intensity values, creating diverse surface patterns that challenge the regression capabilities of SDF models. Each family is parameterised to produce multiple variants, resulting in a rich dataset of 3D shapes with complex intensity distributions.}
\label{fig:data_generation_flow}
\end{figure}

We provide detailed mathematical formulations for all mapper functions used in FPIF. Each mapper function $g:\mathbb{R}\to\mathbb{R}$ converts a signed distance value $d$ to an intensity value. Multiple variants of each family are obtained by varying the parameters.

\noindent{\bf Inverse cube mapper.}
\begin{equation}
g_{\text{inv3}}(d) = \frac{\alpha}{(|d|+\varepsilon)^3},
\end{equation}
where $\alpha>0$ is the intensity scale and $\varepsilon>0$ is a small constant for numerical stability. This produces a sharp intensity peak near the surface ($d=0$) that falls off rapidly with distance.

\noindent{\bf Exponential mapper.}
\begin{equation}
g_{\text{exp}}(d) = \alpha \exp\!\left(-\beta\,\max(-d,0)\right),
\end{equation}
where $\alpha>0$ is the intensity scale and $\beta>0$ controls the decay rate. Intensity decays exponentially from the boundary into the interior.

\noindent{\bf Linear mapper.}
\begin{equation}
g_{\text{lin}}(d) = \max\!\left(0,\alpha-\beta\,\max(-d,0)\right),
\end{equation}
where $\alpha>0$ is the surface intensity and $\beta>0$ is the slope. Intensity decreases linearly with depth, clamped to zero.

\noindent{\bf Floor mapper.}
\begin{equation}
g_{\text{floor}}(d) = \alpha-\delta\left\lfloor \frac{\max(-d,0)}{w}\right\rfloor,
\end{equation}
where $\alpha>0$ is the surface intensity, $w>0$ is the band width, and $\delta>0$ is the per-band intensity step. The interior is partitioned into concentric bands of width $w$; intensity drops by $\delta$ at each band boundary.

\noindent{\bf Modular mapper.}
\begin{equation}
g_{\text{mod}}(d) = \left(\left\lfloor \frac{\max(-d,0)}{w}\right\rfloor \bmod m\right),
\end{equation}
where $w>0$ is the band width and $m\ge 1$ is an integer modulus. Bands repeat with period $m$, producing a layered ring pattern.

\noindent{\bf Sinusoidal mapper.}
\begin{equation}
g_{\text{sin}}(d) = \alpha\sin\!\left(2\pi\,\frac{\max(-d,0)}{\lambda}\right),
\end{equation}
where $\alpha>0$ is the amplitude and $\lambda>0$ is the wavelength. Intensity oscillates sinusoidally as a function of depth from the boundary.

\bibliographystyle{splncs04}
\bibliography{main2}